\def\year{2019}\relax
\documentclass[letterpaper]{article} 
\usepackage{aaai19}  
\usepackage{times}  
\usepackage{helvet}  
\usepackage{courier}  
\usepackage{url}  
\usepackage{graphicx}  
\usepackage{subcaption}
\usepackage{multirow}
\usepackage{color}

\begin{document}
\title{TraffickCam: Explainable Image Matching For Sex Trafficking Investigations}
\author{Abby Stylianou\textsuperscript{1},
Richard Souvenir\textsuperscript{2} and Robert Pless\textsuperscript{3} \\
\textsuperscript{1}Saint Louis University, \textsuperscript{2}Temple University,
\textsuperscript{3}George Washington University\\
abby.stylianou@slu.edu,souvenir@temple.edu,pless@gwu.edu
}
\maketitle
\begin{abstract}
Investigations of sex trafficking sometimes have access to photographs of victims in hotel rooms. These images directly link victims to places, which can help verify where victims have been trafficked or where traffickers might operate in the future.  Current machine learning approaches give promising results in image search to find the matching hotel.  This paper explores approaches to make this end-to-end system better support government and law enforcement requirements, including improved performance, visualization approaches that explain what parts of the image led to a match, and infrastructure to support exporting the results of a query.
\end{abstract}
\section{Introduction}
Modern large-scale image matching approaches offer opportunities to search visual information at scales unimaginable just a few years ago.
It now is possible to automatically search through databases containing billions of images, not just to retrieve generically similar images (e.g. Google Image Search), but also to find and recognize specific people and places. This technology provides some incredible opportunities~\cite{googleLandmarks} and also challenges societal expectations of anonymity and privacy~\cite{oh2016faceless}.

When these technologies are used for official business in the public sector, there is an increasing need to provide explanations of {\em why} machine learning algorithms give the results they do. In the case of image analysis, substantial work has been done to visualize the image regions responsible for classification results, but much less work has explored approaches for image matching.

This paper details work building tools for investigators to determine the hotel where pictures of sex trafficking victims were taken. Our system, TraffickCam, has found that carefully implemented baseline approaches can be effective at this challenging problem~\cite{aipr2015,aipr2017,hotels50k}.
This paper shares work in progress to improve the performance of this image search and visualize what parts of the image are most important for an image match, in order to give investigators greater insight into why the system is suggesting a particular result, and therefore, perhaps, if the system should be trusted.  This supports advice from a recent formal study of how human trafficking investigators interact with modern software support tools~\cite{deeb2019understanding} that concludes: ``When designing tools for law enforcement, it is important to choose algorithms that are human interpretable
and design visualizations that help officers get an intuition for how the process works.''

\begin{figure}
    \centering
    \includegraphics[width=.9\columnwidth]{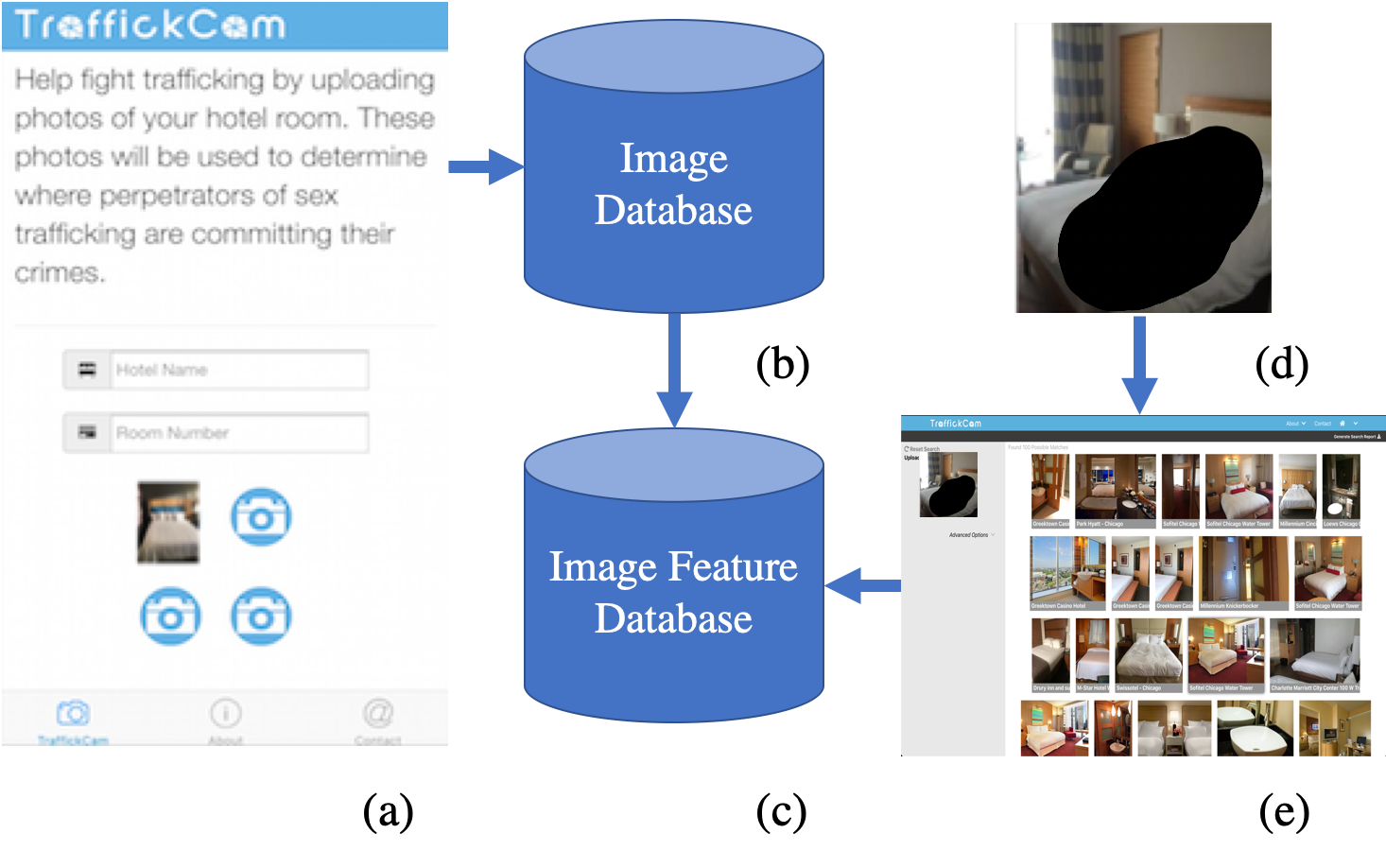}
    \caption{The TraffickCam system consists of a smartphone application (a) to collect relevant imagery of hotel rooms to support investigations of human trafficking. To date, the TraffickCam image database (b) consists of over 3 million hotel room images collected from this application and other publicly available sources of hotel room photos. Deep metric learning is used to convert these images into a searchable index of image features (c). This index supports a law enforcement platform where investigators can upload masked off images from trafficking investigations (d) to retrieve the most similar images in the TraffickCam database (e).}
    \label{fig:frontPage}
\end{figure}
\begin{figure*}[ht]
    \centering
        \begin{subfigure}[b]{.9\columnwidth}
        \includegraphics[width=\columnwidth]{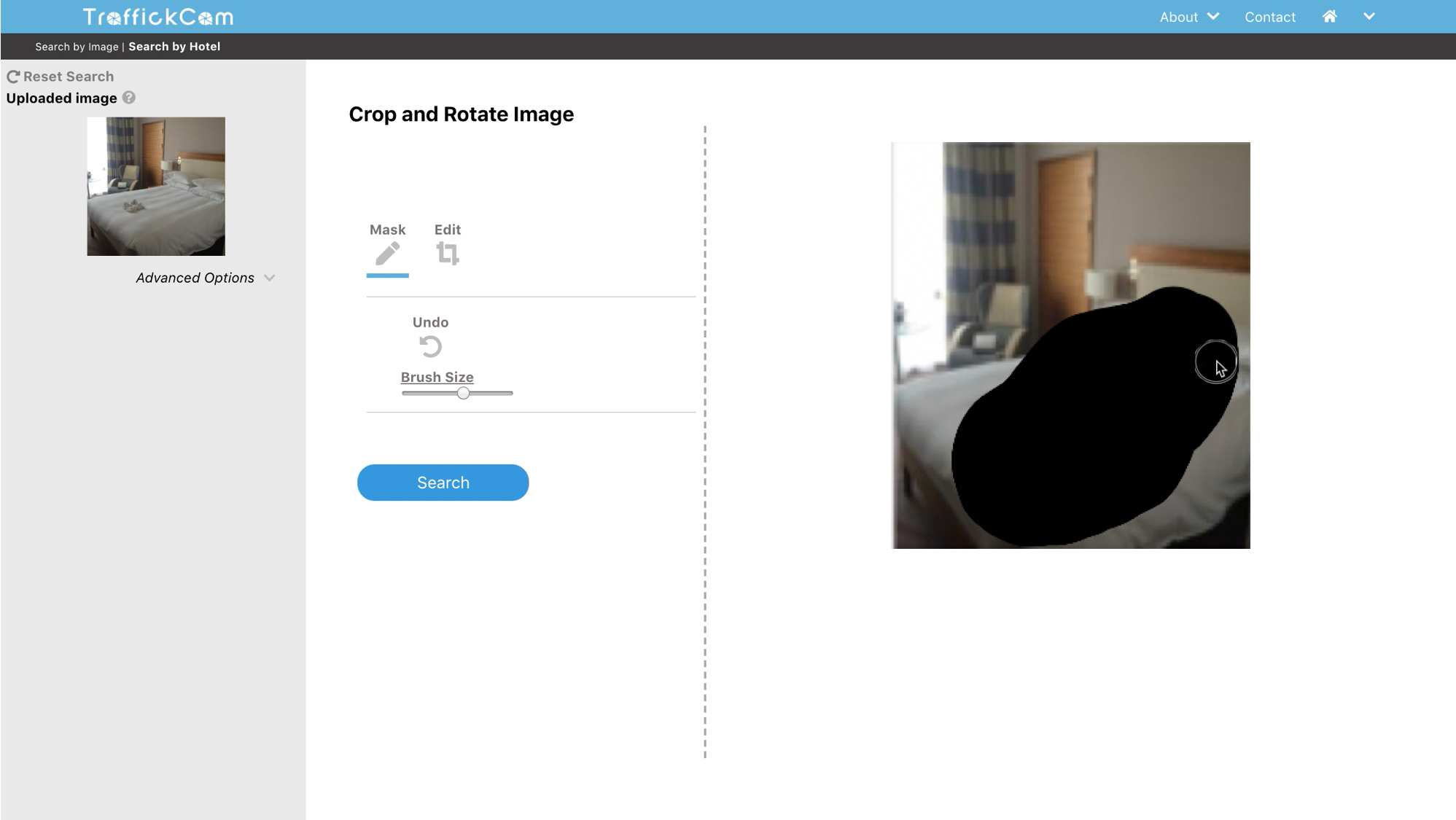}
    \caption{}
    \end{subfigure}
    \\
    \begin{subfigure}[b]{.9\columnwidth}
    \includegraphics[width=\columnwidth]{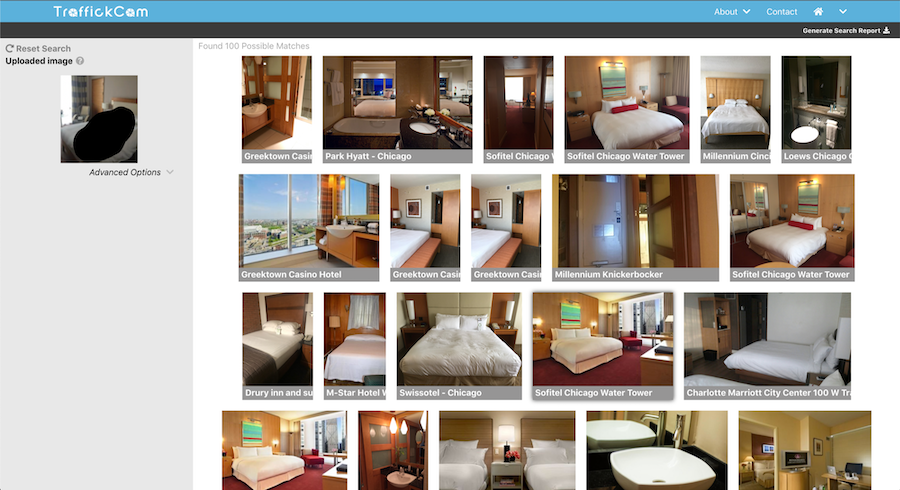}
    \caption{}
    \end{subfigure}
    \begin{subfigure}[b]{.9\columnwidth}
    \includegraphics[width=\columnwidth]{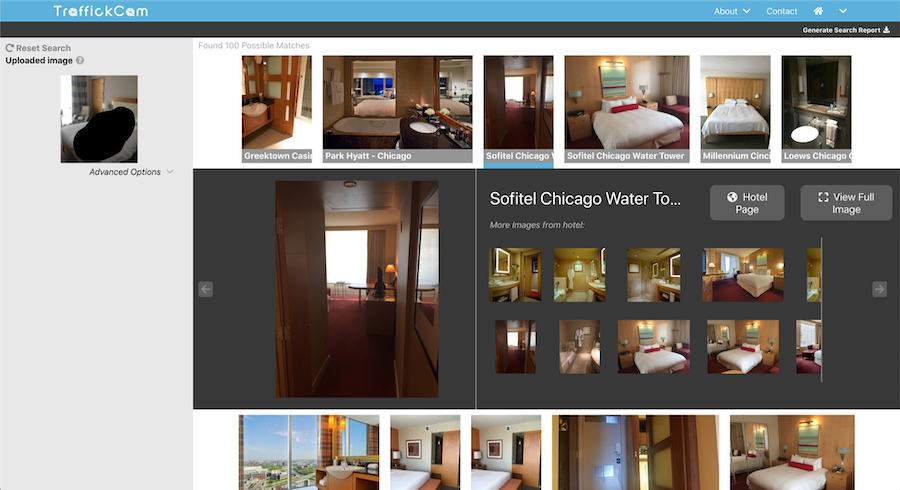}
    \caption{}
    \end{subfigure}
    \caption{(a) The TraffickCam investigator search platform first allows an investigator to mask off any sensitive content in the search image. (b) The search returns the most similar images in the TraffickCam database (searches can be limited by geographic extents or by search terms). (c) Investigators can click on a relevant image to see other images at that hotel. }
    \label{fig:search_resulst}
\end{figure*}

Informal interactions with users of our system and careful examination of the problem domain highlights that learning approaches focused on whole image matching may have limited success in realistic conditions because limited parts of the room may be visible, or, where the database of hotel room images does have examples of similar rooms, they may share some but not all of the same design elements (carpet, lights, headboard, etc.). Therefore, we also explore initial efforts at expanding a visualization tool that highlights the specific elements of the hotel room images match.

We explore the development of search and visualization tools that allow an end-to-end solution to respect the needs of law enforcement and government end users, including the ability to explain and report on why the system is giving the results that it does. Our specific contributions include:

\begin{itemize}
\item updating deep learning training strategies to give better, more interpretable results,
\item improvements to visualization of image matching that highlight important image regions and specifically what regions of images are considered to match, and
\item a search interface infrastructure that supports reporting results for investigations.
\end{itemize}

\section{Related Work}
A limited amount of research has been published on ways to integrate machine learning (ML) to support public sector efforts to fight sex trafficking.  Outside of image analysis, there is text data analysis from the advertising site Backpage~\cite{alvari2017semi} with ML to highlight online advertisements that might be related to sex trafficking, and multi-modal classification of whether online ads are offering paid sex services~\cite{tong2017combating}.  This analysis of individual ads supports work to build knowledge graphs from online escort ads to discover connections that suggest human trafficking based on network properties~\cite{szekely2015building}.

Specific work on image analysis to recognize hotels depicted in advertising images started with local image features~\cite{aipr2015} and progressed to deep learning features~\cite{aipr2017}. Recently, Hotels50k, a public dataset focused on the hotel identification problem was released~\cite{hotels50k}.  In this work, we use this dataset to explore visualization that assist law enforcement in the verification and validation of the image search results.

Visualizations of deep networks for image analysis have focused almost exclusively on classification networks~\cite{zhang2018visual}, highlighting the region of the image most responsible for the classification.  In this paper, we build on one of the few approaches to visualizing similarity networks~\cite{stylianou2019visualizing}, which are the types of networks most commonly used for large-scale image matching.
\newcommand\tabwidth{.71in}
\setlength{\tabcolsep}{1pt}
\renewcommand{\arraystretch}{2}
\begin{figure*}
\centering
\begin{tabular*}{\textwidth}{cccc|ccc|ccc}
     & \multicolumn{3}{c}{Correct Hotel}  & \multicolumn{3}{|c|}{Correct Hotel} & \multicolumn{3}{c}{Incorrect Hotel}  \\
     Query
     & \raisebox{-.5\height}{\includegraphics[width=\tabwidth]{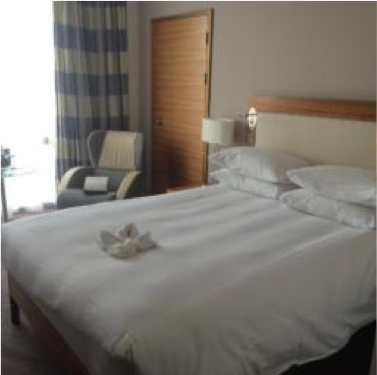}}
     & \raisebox{-.5\height}{\includegraphics[width=\tabwidth]{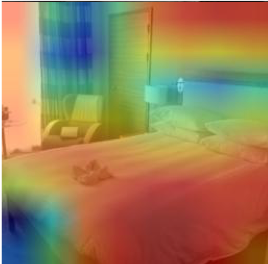}}
     & \raisebox{-.5\height}{\includegraphics[width=\tabwidth]{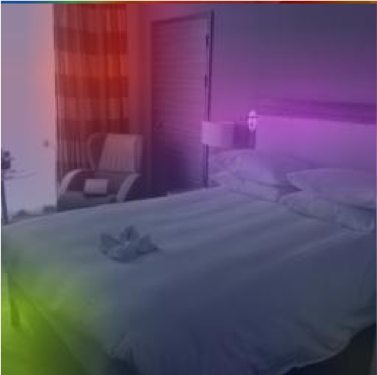}}
     & \raisebox{-.5\height}{\includegraphics[width=\tabwidth]{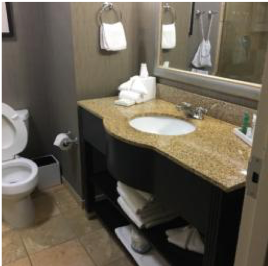}}
     & \raisebox{-.5\height}{\includegraphics[width=\tabwidth]{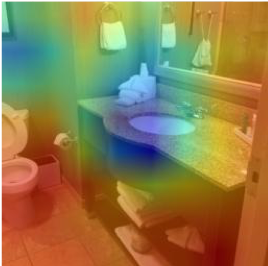}}
     & \raisebox{-.5\height}{\includegraphics[width=\tabwidth]{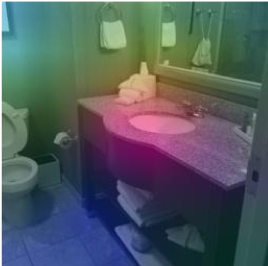}}
     & \raisebox{-.5\height}{\includegraphics[width=\tabwidth]{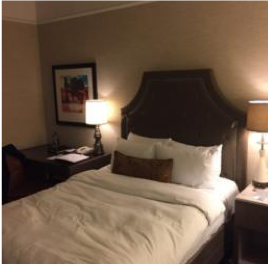}}
     & \raisebox{-.5\height}{\includegraphics[width=\tabwidth]{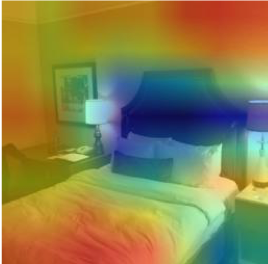}}
     & \raisebox{-.5\height}{\includegraphics[width=\tabwidth]{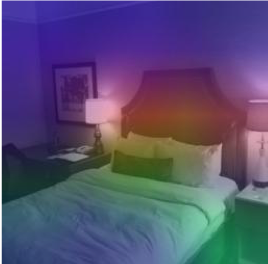}}
     \\
     Result
     & \raisebox{-.5\height}{\includegraphics[width=\tabwidth]{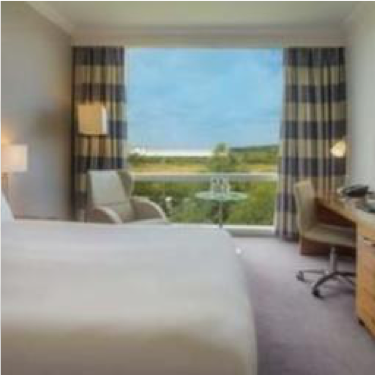}}
     & \raisebox{-.5\height}{\includegraphics[width=\tabwidth]{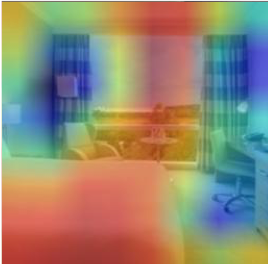}}
     & \raisebox{-.5\height}{\includegraphics[width=\tabwidth]{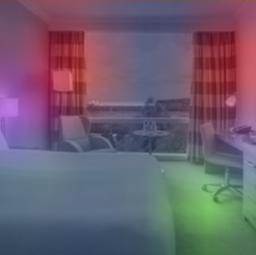}}
     & \raisebox{-.5\height}{\includegraphics[width=\tabwidth]{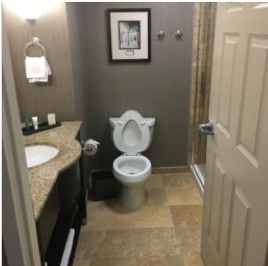}}
     & \raisebox{-.5\height}{\includegraphics[width=\tabwidth]{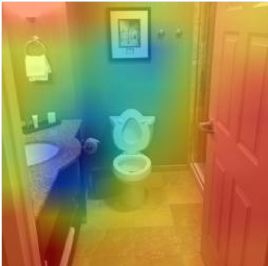}}
     & \raisebox{-.5\height}{\includegraphics[width=\tabwidth]{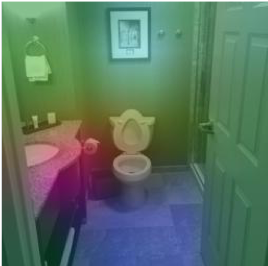}}
     & \raisebox{-.5\height}{\includegraphics[width=\tabwidth]{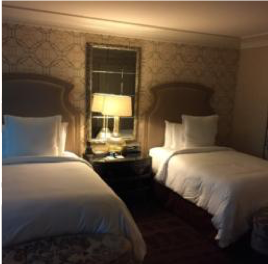}}
     & \raisebox{-.5\height}{\includegraphics[width=\tabwidth]{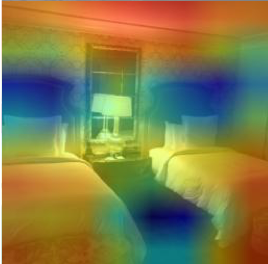}}
     & \raisebox{-.5\height}{\includegraphics[width=\tabwidth]{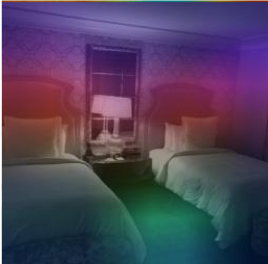}}
     \\
     & (a) & (b) & (c) & (a) & (b) & (c) & (a) & (b) & (c)
\end{tabular*}
\caption{Example visualization results.  Three examples of a query image (top), and the closest matching result image (bottom).  For each, we show (b) figures that highlight which parts of image are most important for the classification (scaled from red to blue where blue is most important), and (c) a PCA based color-coding scheme that colors matching parts of the two images to be the same.}
\label{tab:visualizations}
\end{figure*}

\section{TraffickCam System Design}
The overall system includes a mobile app for crowd-sourcing data (described in ~\cite{aipr2015}), a search engine, a web-front end to support that search, and a report generation system.
We also describe a collection of visualization tools that provide
intuition about system behavior.  These components are modular, which allows for flexible system design and component-wise updates that do not affect the rest of the system. In this section, we review improvements to the underlying search implementation, visualization tools and a system that help investigators generate reports about the search results.

\subsection{Improved Metric Learning for Hotel Matching}
State of the art approaches to large-scale image matching learn deep networks that map images to a feature space.  The training optimizes the network so images from the same hotel are mapped to nearby locations in the feature space and images from different hotels are farther apart.  The hotel matching problem is particularly hard, because multiple rooms from the same hotel may look very different, and even multiple images from the same room may look different (for example if one image includes the bed and the other is of the bathroom).  State of the art results on the Hotels-50K dataset use Resnet-50~\cite{resnet} with careful data augmentation and pre-processing to give about 8\% accuracy in finding the correct hotel as the first result returned by the system (a factor of 4000 improvement over the chance result of 1/50000)~\cite{hotels50k}.

The loss function with which the network was trained was a "batch-all" formulation that forces all images from a single hotel to be mapped to the same location.  Because images from one hotel may look quite different, this requires the network to memorize different viewpoints of the training data and therefore does not generalize as well.  Our best result have come from a refined process for training the network that focuses on "Easy Positive" triplet loss~\cite{xuan2019improved}.  This forces the network to match only the most similar images from a hotel.  This improves the generalization ability and gives about 16\% accuracy in finding the correct hotel as the first response on training data.

\subsection{Improved Visualization Tools}

Resnet-50 is a common deep learning architecture that includes a collection of convolutional layers following by a Global Average Pooling (GAP) layer that converts a spatial feature layout into a single vector to describe the image.  In the current Traffickcam system, this final convolutional layer is $7\times7\times2048$ and the global pooling converts this to a single $2048$ element vector.  This final vector is used to compute the similarity between images.  Previous work~\cite{stylianou2019visualizing} decomposed the final similarity calculation between two $2048$-element vectors to give back $7\times7$ resolution maps of the parts of the image that were most important for the similarity assessment.

Figure~\ref{tab:visualizations} shows example queries (top) and results that were shown to be similar (bottom).  Visualization of which parts of an image results led them to be judged similar are shown labelled as (b), with the most important regions shown in blue. This gives a general assessment of which parts of the image are important, but does not specify specifically which components of the query image correspond to particular components of the result image.

In Figure~\ref{tab:visualizations}, the color coded images labeled (c) are early results from a visualization approach that attempts to gain more insight about the specific relationship between images. To construct these visualizations for a pair of images that are similar, we consider the $7\times7\times2048$ convolutional layer as 49 vectors (each of length $2048$) describing the local features of each image. These 49 vectors from each image are concatenated to get 98 total vectors.  We can then run PCA on these vectors to get principle components (each of size $2048$) that describe patterns of how the the local image description vary within and between the images.  We then display the top 3 components as an RGB color mapped back to the locations on the image pair. This visualization has the property that similarly colored locations in the visualization actually have similar representations in the feature space, and therefore correspond to each other.

On the left set of results from Figure~\ref{tab:visualizations}, this visualization approach shows that the network is matching the curtains (because both images have curtains highlighted in red), the floor (in green), and the headboard/wall light (in purple).  The center image matches the floors, the sink, but not, for example, the toilet. The third set of results, on the right, are from a pair of images with a high similarity score but from different hotels. The PCA-based visualization approach highlights the similar headboard in red, but incorrectly shows that the network is mapping the linens and pillows on the bed in the query image to the carpet in the result image (highlighted in green).

\subsection{Automatic Report Generation}
\begin{figure}
    \centering
    \includegraphics[width=\columnwidth]{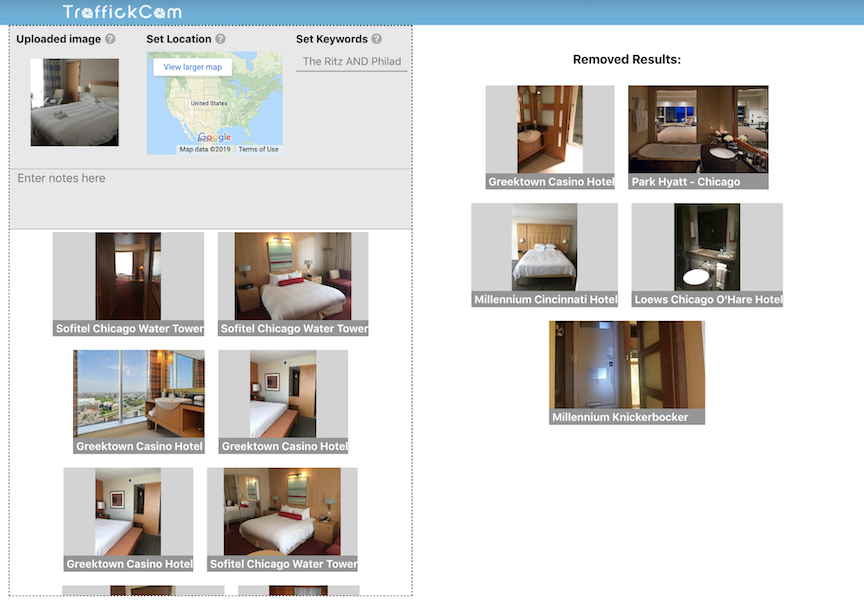}
    \caption{Investigators can select particular results from a query to include in a summary report. This report contains the masked query image, search criteria, any notes that the investigator adds, and the selected results. This report can be saved or printed directly from the website.}
    \label{fig:report}
\end{figure}

In the context of an investigative process, the result of using such a system is a report
to be shared with law enforcement agents or other stakeholders who may not be knowledgeable in machine learning or image retrieval. Thus, it is important that the final report is self-contained and convincing. Traffickcam provides functionality for summarizing and curating the results of a query. Figure~\ref{fig:report} shows the automated
report generation page, which includes the top matching images, information on the most likely
hotel matches, and a space for the analyst to include notes. Additionally, the analyst has the ability to add, remove, or re-arrange the matching images returned by the automated system in order to frame the results in a way that best supports the conclusions and/or conveys the appropriate level of certainty.

\section{Conclusion}
TraffickCam is an existing system to support sex trafficking investigations by identifying the hotel where an image was taken.  It is currently in use at the National Center for Missing and Exploited Children.  This report details work in progress to improve this system both in terms of system accuracy and to make it better support the public sector use cases through improved results, and increased explainability of those results.

One interesting issue is that evaluating search results in this domain is not necessarily best done by comparing results to ground truth (i.e., the evaluation metric proposed in~\cite{hotels50k}).  The primary use of the system is to suggest possible hotel matches, which are then evaluated by an investigator.  If the system matches the query to the correct hotel, but shows views of that hotel where the matching features are not obvious to the investigator, they are likely to miss this match, even though the automated part of the search was effective.  Finding ways to integrate visualization tools within the search results and evaluating the end-to-end system performance are important next steps.

\small\bibliography{bib}

\begin{thebibliography}{}

\bibitem[\protect\citeauthoryear{Alvari, Shakarian, and
  Snyder}{2017}]{alvari2017semi}
Alvari, H.; Shakarian, P.; and Snyder, J.~K.
\newblock 2017.
\newblock Semi-supervised learning for detecting human trafficking.
\newblock {\em Security Informatics} 6(1):1.

\bibitem[\protect\citeauthoryear{Deeb-Swihart, Endert, and
  Bruckman}{2019}]{deeb2019understanding}
Deeb-Swihart, J.; Endert, A.; and Bruckman, A.
\newblock 2019.
\newblock Understanding law enforcement strategies and needs for combating
  human trafficking.
\newblock In {\em Proceedings of the 2019 CHI Conference on Human Factors in
  Computing Systems},  331.
\newblock ACM.

\bibitem[\protect\citeauthoryear{He \bgroup et al\mbox.\egroup }{2015}]{resnet}
He, K.; Zhang, X.; Ren, S.; and Sun, J.
\newblock 2015.
\newblock Deep residual learning for image recognition.
\newblock {\em CoRR} abs/1512.03385.

\bibitem[\protect\citeauthoryear{Oh \bgroup et al\mbox.\egroup
  }{2016}]{oh2016faceless}
Oh, S.~J.; Benenson, R.; Fritz, M.; and Schiele, B.
\newblock 2016.
\newblock Faceless person recognition: Privacy implications in social media.
\newblock In {\em European Conference on Computer Vision},  19--35.
\newblock Springer.

\bibitem[\protect\citeauthoryear{Stylianou \bgroup et al\mbox.\egroup
  }{2015}]{aipr2015}
Stylianou, A.; Norling-Ruggles, A.; Souvenir, R.; and Pless, R.
\newblock 2015.
\newblock Indexing open imagery to create tools to fight sex trafficking.
\newblock {\em 2015 IEEE Applied Imagery Pattern Recognition Workshop (AIPR)}
  00:1--6.

\bibitem[\protect\citeauthoryear{Stylianou \bgroup et al\mbox.\egroup
  }{2019}]{hotels50k}
Stylianou, A.; Xuan, H.; Shende, M.; Souvenir, R.; and Pless, R.
\newblock 2019.
\newblock Hotels-50k: A global hotel recognition dataset.
\newblock In {\em AAAI}.

\bibitem[\protect\citeauthoryear{Stylianou, Souvenir, and
  Pless}{2017}]{aipr2017}
Stylianou, A.; Souvenir, R.; and Pless, R.
\newblock 2017.
\newblock Traffickcam: Crowdsourced and computer vision-based approaches to
  fighting sex trafficking.
\newblock {\em 2017 IEEE Applied Imagery Pattern Recognition Workshop (AIPR)}.

\bibitem[\protect\citeauthoryear{Stylianou, Souvenir, and
  Pless}{2019}]{stylianou2019visualizing}
Stylianou, A.; Souvenir, R.; and Pless, R.
\newblock 2019.
\newblock Visualizing deep similarity networks.
\newblock In {\em 2019 IEEE Winter Conference on Applications of Computer
  Vision (WACV)},  2029--2037.
\newblock IEEE.

\bibitem[\protect\citeauthoryear{Szekely \bgroup et al\mbox.\egroup
  }{2015}]{szekely2015building}
Szekely, P.; Knoblock, C.~A.; Slepicka, J.; Philpot, A.; Singh, A.; Yin, C.;
  Kapoor, D.; Natarajan, P.; Marcu, D.; Knight, K.; et~al.
\newblock 2015.
\newblock Building and using a knowledge graph to combat human trafficking.
\newblock In {\em International Semantic Web Conference},  205--221.
\newblock Springer.

\bibitem[\protect\citeauthoryear{Tong \bgroup et al\mbox.\egroup
  }{2017}]{tong2017combating}
Tong, E.; Zadeh, A.; Jones, C.; and Morency, L.-P.
\newblock 2017.
\newblock Combating human trafficking with deep multimodal models.
\newblock {\em arXiv preprint arXiv:1705.02735}.

\bibitem[\protect\citeauthoryear{Xuan, Stylianou, and
  Pless}{2019}]{xuan2019improved}
Xuan, H.; Stylianou, A.; and Pless, R.
\newblock 2019.
\newblock Improved embeddings with easy positive triplet mining.
\newblock {\em arXiv preprint arXiv:1904.04370}.

\bibitem[\protect\citeauthoryear{Zhang and Zhu}{2018}]{zhang2018visual}
Zhang, Q.-s., and Zhu, S.-C.
\newblock 2018.
\newblock Visual interpretability for deep learning: a survey.
\newblock {\em Frontiers of Information Technology \& Electronic Engineering}
  19(1):27--39.

\bibitem[\protect\citeauthoryear{Zheng \bgroup et al\mbox.\egroup }{June
  2009}]{googleLandmarks}
Zheng, Y.-T.; Zhao, M.; Song, Y.; Adam, H.; Buddemeier, U.; Bissacco, A.;
  Brucher, F.; Chua, T.-S.; and Neven, H.
\newblock June, 2009.
\newblock Tour the world: building a web-scale landmark recognition engine.
\newblock In {\em IEEE Conference on Computer Vision and Pattern Recognition}.

\end{thebibliography}
\bibliographystyle{aaai}
\end{document}